\documentclass[acmtog]{acmart}
\makeatletter
\let\@authorsaddresses\@empty
\makeatother

\usepackage{soul}
\usepackage{booktabs} 
\setcitestyle{square}
\usepackage{ifthen}
\usepackage{enumitem}
\usepackage{algorithm}
\usepackage[noend]{algpseudocode}
\usepackage{float}
\usepackage{syntax}
\usepackage{amsfonts}

\usepackage{listings}
\usepackage{fancyvrb}
\usepackage{wrapfig}
\usepackage{graphicx}
\usepackage{subfigure}
\usepackage{xspace}
\usepackage{colortbl}
\usepackage{gensymb}
\usepackage{verbatim}
\usepackage{colortbl}
\usepackage{booktabs}
\usepackage{multirow}
\usepackage{cleveref}
\usepackage{pifont}
\usepackage{circledsteps}

\usepackage{amsthm,amsmath,amssymb}
\usepackage{mathrsfs}
\usepackage[normalem]{ulem}
\begin{document}
\settopmatter{printacmref=false}
\setcopyright{none}
\renewcommand\footnotetextcopyrightpermission[1]{}
\pagestyle{empty} 

\title{SymBridge: A Human-in-the-Loop Cyber-Physical Interactive System for Adaptive Human-Robot Symbiosis}

\author{Haoran Chen}
\authornote{Joint first authors.}
\affiliation{%
  \institution{Shandong University}
  \country{China}
}
\author{Yiteng Xu}
\authornotemark[1]
\affiliation{%
  \institution{ShanghaiTech University}
  \country{China}
}
\author{Yiming Ren, Yaoqin Ye}
\affiliation{%
  \institution{ShanghaiTech University}
  \country{China}
}

\author{Xinran Li, Ning Ding, Yuxuan Wu, Yaoze Liu}
\affiliation{%
  \institution{Shandong University}
  \country{China}
}

\author{Peishan Cong, Ziyi Wang, Bushi Liu, Yuhan Chen}
\affiliation{%
  \institution{ShanghaiTech University}
  \country{China}
}
\author{Zhiyang Dou}
\affiliation{%
  \institution{The University of Hong Kong}
  \country{China}
}
\author{Xiaokun Leng}
\affiliation{%
  \institution{LEJU(Shenzhen) Robotics Co., Ltd}
  \country{China}
}
\author{Manyi Li}
\authornote{Corresponding authors.}
\affiliation{%
  \institution{Shandong University}
  \country{China}
}
\author{Yuexin Ma}
\authornotemark[2]
\affiliation{%
  \institution{ShanghaiTech University}
  \country{China}
}
\author{Changhe Tu}
\affiliation{%
  \institution{Shandong University}
  \country{China}
}

\begin{abstract}
The development of intelligent robots seeks to seamlessly integrate them into the human world, providing assistance and companionship in daily life and work, with the ultimate goal of achieving human-robot symbiosis. This requires robots with intelligent interaction abilities to work naturally and effectively with humans.
However, current robotic simulators fail to support real human participation, limiting their ability to provide authentic interaction experiences and gather valuable human feedback essential for enhancing robotic capabilities.
In this paper, we introduce \textbf{SymBridge}, the first human-in-the-loop cyber-physical interactive system designed to enable the safe and efficient development, evaluation, and optimization of human-robot interaction methods. Specifically, we employ augmented reality technology to enable real humans to interact with virtual robots in physical environments, creating an authentic interactive experience. Building on this, we propose a novel robotic interaction model that generates responsive, precise robot actions in real time through continuous human behavior observation. The model incorporates multi-resolution human motion features and environmental affordances, ensuring contextually adaptive robotic responses. Additionally, SymBridge enables continuous robot learning by collecting human feedback and dynamically adapting the robotic interaction model. 
By leveraging a carefully designed system architecture and modules, \textit{SymBridge} builds a bridge between humans and robots, as well as between cyber and physical spaces, providing a natural and realistic online interaction experience while facilitating the continuous evolution of robotic intelligence. Extensive experiments, user studies, and real robot testing demonstrate the system's promising performance and highlight its potential to significantly advance research on human-robot symbiosis.
\end{abstract}




\begin{teaserfigure}
  \includegraphics[width=1.0\linewidth]{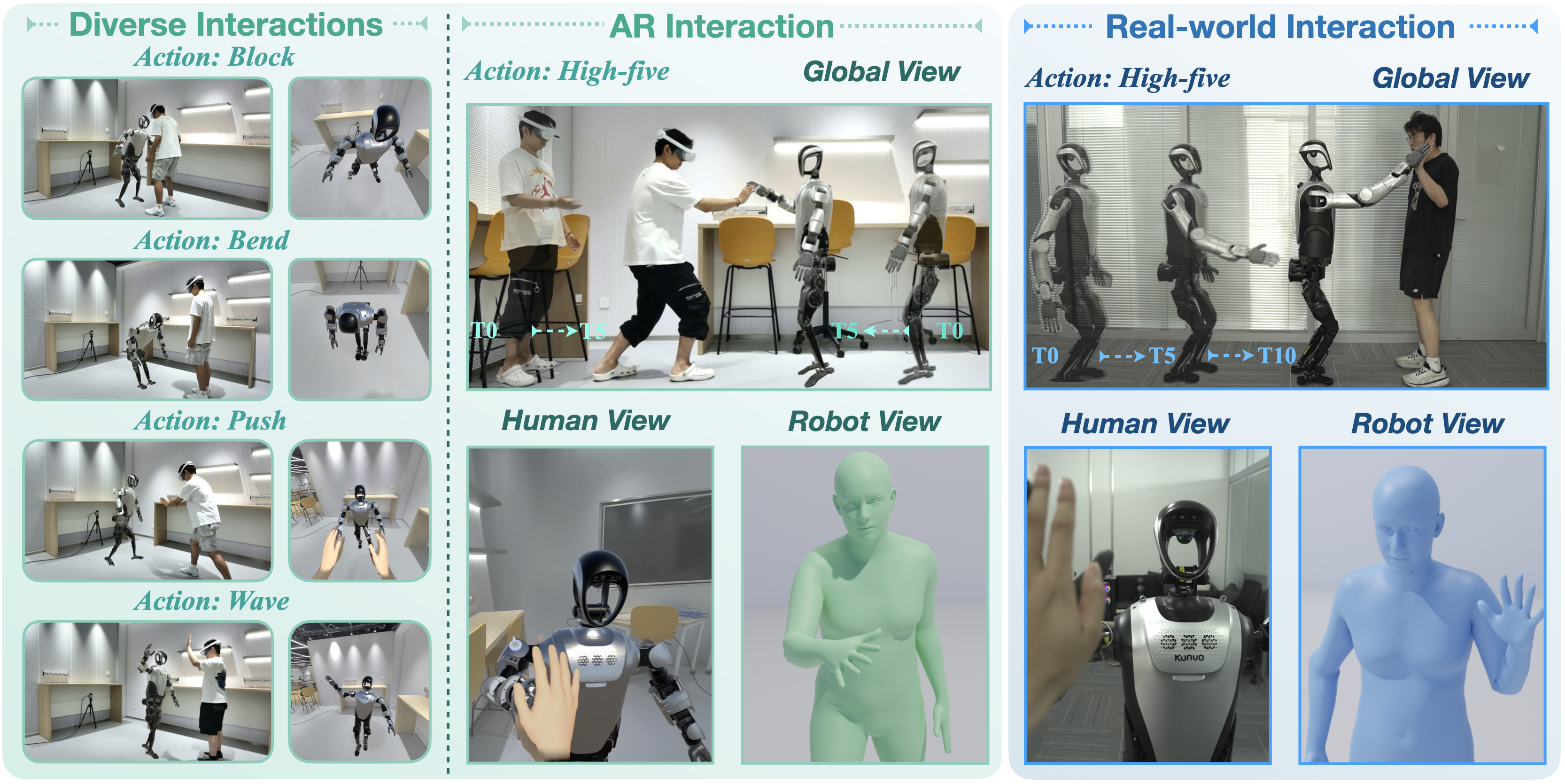}
	\vspace{-15pt}
\caption {The SymBridge system enables natural real human-virtual robot interaction via an AR interface, supporting diverse actions like high-fives, pushes, and waves. Its interactive model improves continuously through human feedback and can eventually deploy to real robots for real-world use. The left image demonstrates SymBridge’s cyber-physical interaction with diverse interaction categories, and the right displays real-world human-robot interaction.}
  \label{fig:teaser}    
\end{teaserfigure}

\maketitle

\section{Introduction}
\label{sec:intro}

With rapid advancements in robotics and artificial intelligence, robots are moving from industrial settings into everyday human environments~\cite{lecun2022path,gonzalez2021service}, showing promise in areas like daily service, home care, and healthcare. To achieve the vision of human-robot symbiosis, the most challenging issue is how to enable robots to interact and collaborate with humans as naturally and efficiently as possible in daily life.

To support the development and evaluation of intelligent robotic algorithms, over the past few decades, many robotic simulators~\cite{makoviychuk2021isaac, todorov2012mujoco,igibson, nasiriany2024robocasa} have emerged and been widely used. They are well-suited for tasks such as robotic navigation, manipulation, and control within predefined digital environments containing static objects. However, they struggle to support human-robot interaction (HRI), as modeling human behavior—especially the complex, dynamic, and closed-loop interactive behavior—is extremely challenging. To bridge this gap, recent work~\cite{HumanTHOR, liu2024collabsphere} has shifted toward incorporating real humans into HRI simulations. However, these systems still have limitations: humans generally control virtual avatars through keyboards or VR interfaces to collaborate with simulated robots. The focus is on completing tasks that do not involve close physical interaction. So far, there still lacks 3D physical interaction, preventing truly realistic and immersive human-robot behavioral exchange. 

To address this gap, our system aims to enable real-time and realistic human–robot interaction through AR. AR allows humans to interact directly with virtual robots, and through repeated interactions, users develop familiarity with robot behaviors. Over time, the system collects human feedback that continuously improves robot motion generation. Together, these elements represent a first step toward achieving human–robot symbiosis.

To facilitate the development, evaluation, and evolution 
of HRI algorithms—ultimately enabling intelligent robots to provide direct services to humans—a simulation system supporting authentic HRI experiences is crucial. However, this presents significant challenges: 
\textbf{1)} How to involve real humans interacting with virtual robots in physical space for authentic interaction experience? \textbf{2)} How to drive virtual robots to respond to human actions in real-time, achieving natural and seamless interactive behaviors? \textbf{3)} How to enable continuous algorithmic self-improvement through ongoing interaction, ultimately creating intelligent robots that meet human expectations?

In this paper, we present SymBridge, a pioneering human-in-the-loop robotic interaction system designed for achieving human-robot symbiosis, as Fig.~\ref{fig:teaser} shows. 
SymBridge innovatively addresses the above challenges with the following key characteristics: 

\noindent\textit{\textbf{1) Realistic Interaction Interface}}: SymBridge simulates robot perception with a LiDAR-based motion capture system and presents robot reactions through augmented reality (AR) technology. The LiDAR-based MoCap module implements real-time 3D human motion digitization while robust to varying lighting conditions and ensuring privacy-preserving. The AR interface bridges the physical and virtual realms by rendering virtual robots into the user’s environment via AR glasses, creating an immersive, authentic interactive experience that seamlessly blends human and robotic actions. 

\noindent\textit{\textbf{2) Real-Time Robotic Interaction Model}}:
Unlike offline motion generation methods, SymBridge's interaction model enables online, real-time robot action generation. Its key innovation lies in the synergistic integration of spatial reasoning and temporal understanding. Specifically, it fuses an Affordance Predictor for fine-grained spatial awareness with a Multi-Resolution Human Feature (MRHF) Learner that captures deep temporal context. This holistic approach, distinct from prior work (e.g., JRT~\cite{xu2023joint}, ReGenNet~\cite{xu2024regennet}) that often addresses these aspects separately, allows the model to generate responses that are simultaneously contextually and socially-aware, adapting instantly to rapid changes in human behavior.

\noindent\textit{\textbf{3) Feedback-Driven Adaptation Mechanism}}:
SymBridge continuously collects human feedback during interactions and uses this data to iteratively fine-tune the robot’s capabilities. This creates a "better with use" evolution mechanism, where the robot’s behavior progressively aligns with user preferences, significantly simplifying and boosting the improvement of robotic intelligence.

\noindent Together, these components establish SymBridge as a pioneering system for human-robot interaction, 
which enables immersive, intuitive, and efficient interaction paradigms.
As a dual bridge between humans and robots and physical and virtual worlds, SymBridge accelerates the realization of human-robot symbiosis, with the potential to pave the way for collaborative ecosystems where robots and humans evolve in tandem.

To demonstrate the effectiveness of our solution, we have developed a comprehensive evaluation methodology and benchmark for human-robot interaction, complementing both quantitative and qualitative assessments of the algorithmic components within our system.
Further, we evaluate the efficiency and quality performance of our system when facilitating human users to interact with the virtual robot. 
A user empirical test with 50 participants confirms that human users are satisfied with the realistic and real-time interaction experience via our system. We also deploy the generated virtual robot reactions to a real humanoid robot to examine the realism and fidelity of the robot movements presented with our system. Additionally, by fine-tuning our model with user feedback data, we demonstrate continuous performance improvement, proving our system's capability to enhance robotic interaction models through ongoing interactive data collection.

\section{Related Work}
\label{sec:related}

\subsection{Robotic Simulators}
Simulators play a key role in robotics research by enabling safe, cost-effective testing and rapid development across diverse scenarios. Physical simulators~\cite{coumans2021,todorov2012mujoco,makoviychuk2021isaac} primarily focus on simulating the motion control, kinematics, and dynamics of robots. They are critical for the reinforcement learning for manipulation and control tasks.
To advance the study of embodied intelligence in 3D environments, a growing number of robotic simulators~\cite{kolve2017ai2, deitke2020robothor, ehsani2021manipulathor,shen2021igibson, li2021igibson, li2023behavior,savva2019habitat, szot2021habitat, puig2023habitat,nasiriany2024robocasa,wang2024grutopia} have emerged. 
These platforms 
can benefit tasks such as robotic perception~\cite{wang2024embodiedscan}, navigation~\cite{vuong2024habicrowd}, and manipulation~\cite{miller2004graspit}. 
However, these simulators only focus on passive environment without active characters and do not support human-robot interaction simulations, which are essential for service robot applications. They also exclude direct human participation, preventing true human-involved simulations.


\subsection{Human-robot Interaction}
As robots become increasingly integrated into everyday life and various industries, improving how humans and robots interact, collaborate, and coexist is critical to enhancing the effectiveness, safety, and social acceptance of robotic systems~\cite{cakmak2011human}. Recent studies have explored the use of virtual reality (VR) or augmented reality (AR) devices~\cite{chen2024arcap, yang2024arcade, park2024dexhub, nechyporenko2024armada} and developed teleoperation systems~\cite{wonsick2021human,mosbach2022accelerating, iyer2024open, cheng2024open} to capture high-quality robot data. However, these approaches primarily position humans as instructors, either by providing task instructions or performing demonstrations for the robots—rather than focusing on true human-robot interaction. ~\cite{HumanTHOR,liu2024collabsphere} introduce approaches, enabling humans to use VR devices or motion capture system to drive digital avatars to interact with robots.
However, they inherently restrict humans from engaging with the robot in the real physical environment, as opposed to through a virtual avatar, creating a disconnection between human intentions and robotic responses. Moreover, the absence of real-time, authentic human feedback hinders the robot’s ability to iteratively learn, adapt, and refine its behaviors.

It is worth distinguishing our work from traditional collaborative robots (cobots). While cobots typically focus on task-oriented collaboration in structured industrial settings (e.g., assembly lines)~\cite{colgate1996cobots,el2019cobot,hentout2019human}, SymBridge targets open-ended, social interactions in service environments (e.g., companionship, assistance). Our key focus is on learning from full-body human dynamics to enable flexible and natural interaction beyond purely work-oriented tasks.

\subsection{Interactive Motion Generation} 
Human behavior is inherently uncertain and highly dynamic, making human-robot interaction and collaboration particularly challenging. Traditional motion planning algorithms~\cite{dai2014whole} struggle to achieve real-time optimization in high-degree-of-freedom scenarios and RL-based methods~\cite{hwangbo2019learning} lack the generalization capability for complex behaviors. Our work leverages deep learning-based generative methods and human-human interaction data to learn human-humanoid interaction behaviors.
Previous research on human-interactive generation has primarily focused on static environment~\cite{jiang2022chairs,kulkarni2024nifty,jiang2023full}, often neglecting the dynamic aspects of interactions.
With the emergence of several HOI and HHI datasets~\cite{li2023object,bhatnagar2022behave,liang2024intergen}, some works have attempted to address dynamic object or human-human interactions.
They either generate human state condition on the entire given interactive subject sequence~\cite{li2023object,xu2024regennet,cong2025semgeomo} or predefined trajectory~\cite{li2023controllable}, or jointly generate both object and human states based on textual descriptions~\cite{peng2023hoi,diller2024cg,wu2024thor,xu2023interdiff}. 
However, for natural interaction, the robot is required to generate responsive motions that correspond to dynamic interactions in real time. 
InterDiff~\cite{xu2023interdiff} 
predicts the next step in interactions while suffering from slow processing speeds, limiting their practicality for real-time scenarios. 
To address these limitations, we propose an online high-quality robot motion generation, which enables the robot to dynamically respond to rapid changes in interactive behavior.

\begin{figure*}[t]
\centering
\includegraphics[width=1\linewidth]{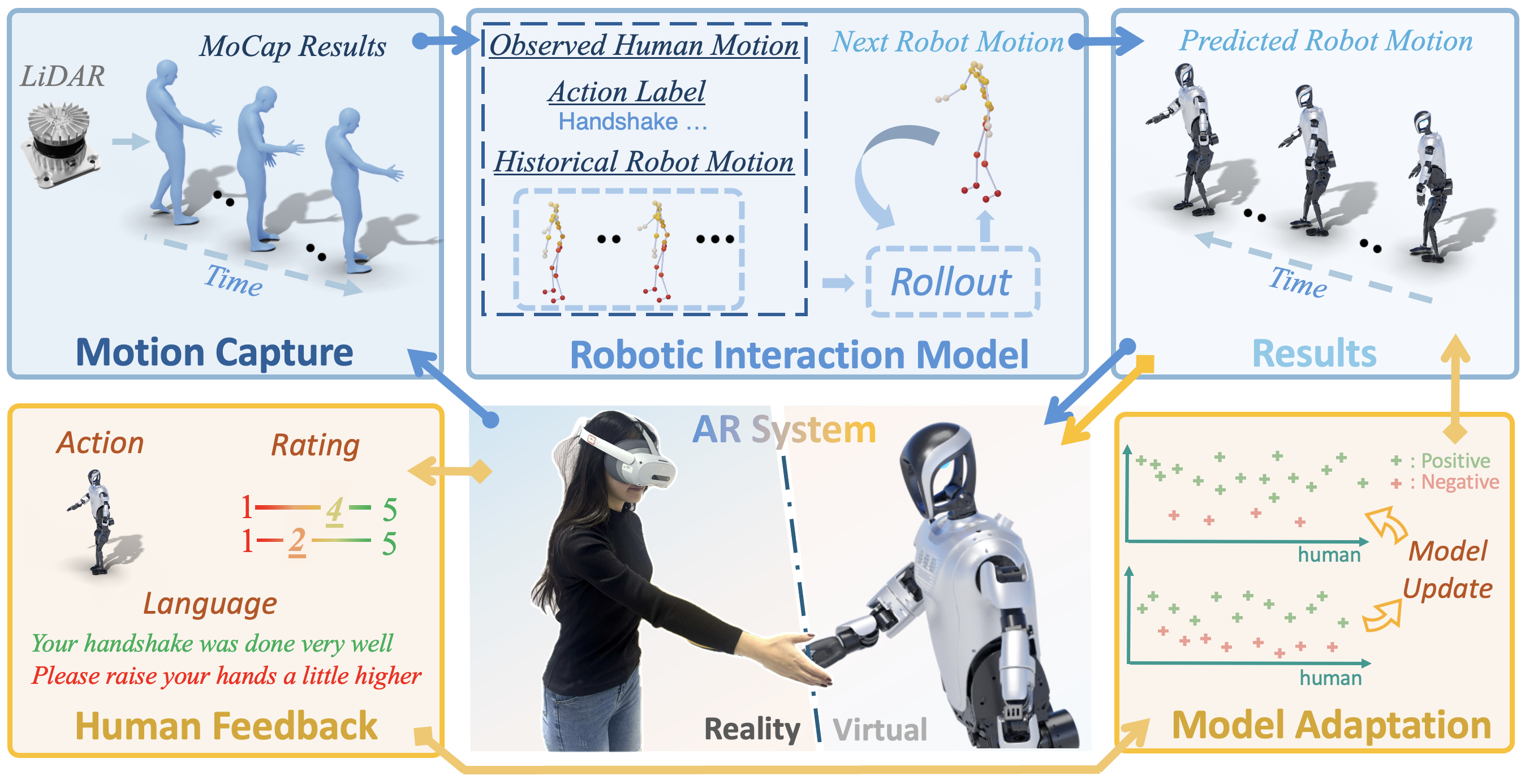}
\caption{System overview of SymBridge. Humans interact with virtual robots in 3D physical space through AR technology with a strong sense of realism. The robot perceives human actions via motion capture and generates corresponding responses through the robotic interaction model. Through the blue cycle, humans can repeatedly engage with the robot, refining their understanding and trust for robots' behavior. Meanwhile, leveraging the human-in-the-loop advantage, human feedback is collected and used to identify valuable data for robotic model adaptation. The yellow cycle enables the robot to continuously enhance its interactive capability based on human feedback, progressively aligning its actions with human preferences.}
\label{fig:system}
\end{figure*}

\begin{figure*}
\centering
\includegraphics[width=1\linewidth]{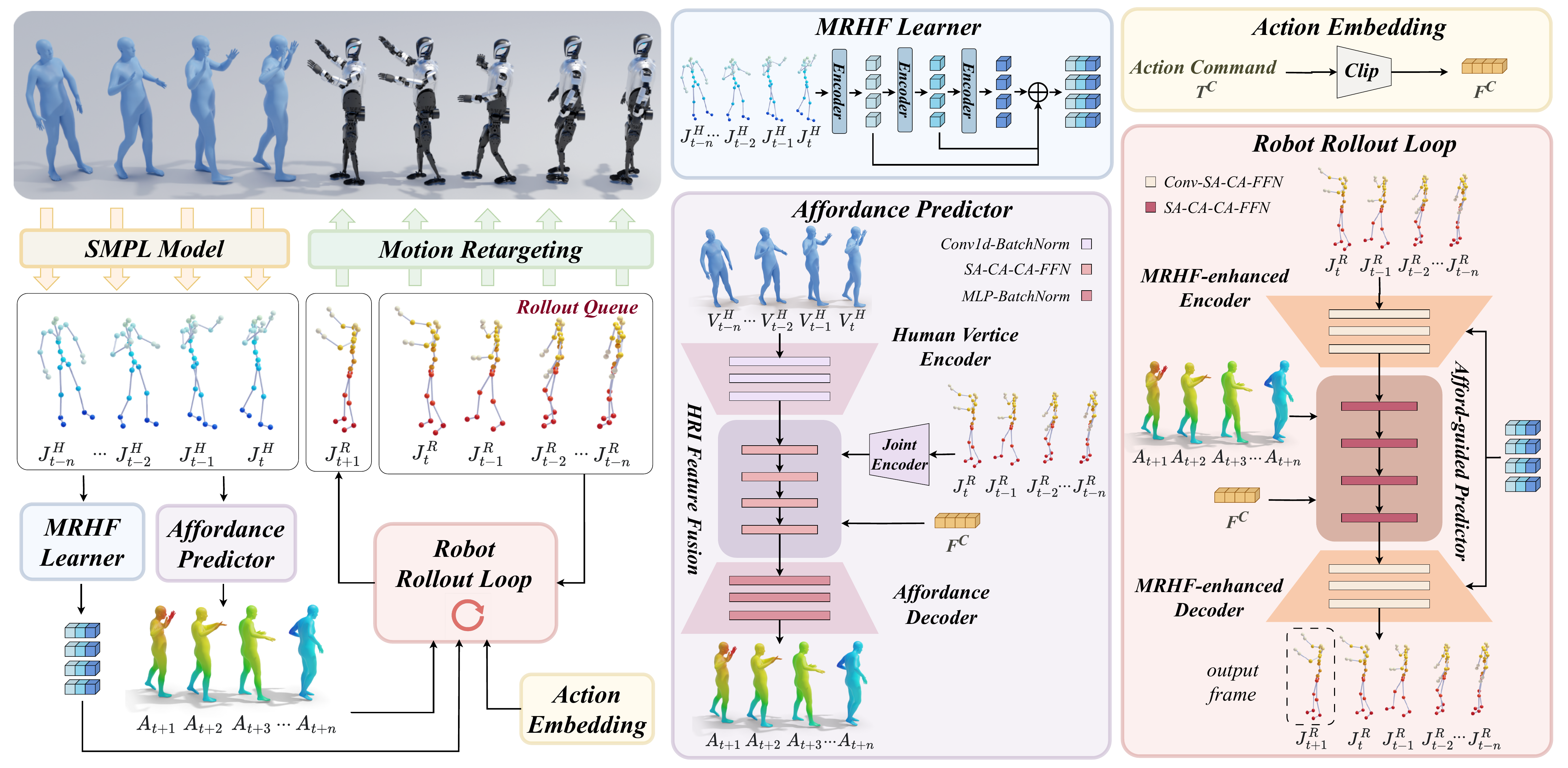}
\vspace{-20pt}
\caption{Overall pipeline of our Robotic Interaction Model. The model ingests text action commands, historical robot joints, observed human joints and SMPL mesh vertices, and processes them through four tightly coupled modules. The Affordance Predictor explicitly estimates dense end-effector–vertex spatial fields to enable dexterous, socially-appropriate contact reasoning; the Multi-Resolution Human Feature (MRHF) Learner fuses hierarchical temporal cues to boost situational awareness; the Robot Rollout Loop leverages these affordance and human features to iteratively generate coherent, context-adaptive joint trajectories for real-time responsiveness; and finally, the Motion Retargeting module maps each predicted skeleton frame to robot-specific angle axis.}
\label{fig:interactiveModel}
\vspace{-10pt}
\end{figure*}
\section{Design Goals}

Towards the human-robot symbiosis, our primary purpose is to develop a human-in-the-loop system that supports the versatile interactions between humans and robots, such as hand-shaking, hugging, etc. In this way, human users can easily obtain more experiences to get used to interacting with robots and returning their authentic feedback to improve robot skills to better facilitate human life. We decompose it into specific design goals as follows.

\noindent\textbf{D1. Simulate a realistic and real-time human-robot interaction experience.} To allow human users to interact with virtual robots, the system must capture human motion, produce a robot reaction, and present it to the user. More importantly, the whole pipeline should be realistic and real-time enough for the user to have an immersive experience like interacting in the real world.

\noindent\textbf{D2. Produce reasonable and plausible robot reactions regarding human behavior.} 
Humans may perform various actions during interactions. The fundamental capability of the system is to automatically determine the appropriate robot reaction based on human behavior to ensure a fluent human-robot interaction.

\noindent\textbf{D3. Facilitate human feedback collection for robot interaction skill enhancement.} Over the long-term interactions, humans and robots should mutually adapt to each other to achieve harmonious co-existence. It requires an efficient fine-tuning mechanism to continuously enhance the robot's interaction skills. Therefore, the system should facilitate the collection of human-robot interaction data with authentic human feedback and efficiently utilize the collected data to fine-tune robot interactions.

\section{Symbridge System}
\subsection{System Architecture}

The aforementioned design goals lead to our system architecture, as illustrated in Fig.~\ref{fig:system}. The core is the immersive human-robot interaction with our AR interface, where a human user can wear AR glasses to interact with the virtual robot in the real environment. It activates the two main capabilities of our system.

The \emph{human-robot interaction}, illustrated with the blue arrows in Fig.~\ref{fig:system}, aims to support the realistic and real-time interactions between humans and the virtual robot. First, the human behavior is continuously acquired as motion sequences by the motion capture module. Given the current frame and the cached historical frames of the captured human motion, we employ an interactive model to predict the next frames of the robot motion. The predicted robot motion is sent 
to the applications running on the AR glasses to provide immersive interaction experiences. Our system also provides a third-person perspective by presenting the captured real human and rendered virtual robot on the same visual frame.

The \emph{model adaptation}, illustrated with the yellow arrows in Fig.~\ref{fig:system}, aims to enhance the robot's interaction skills by collecting and utilizing the human-in-the-loop interaction data. During the human-robot interaction through the AR interface, the human users may express their feedback, such as ratings on the quality of interaction experiences. Our system records the interaction process as well as the human feedback, and utilizes them to fine-tune the interactive model to adapt it to human preference. 

\subsection{Modules for Human-Robot Interaction}
\label{sec:mocap}

\noindent\textbf{Motion Capture.} To enable realistic and real-time human–robot interaction, the system requires accurate motion capture that approximates the robot’s perception. We employ an Ouster OS1-128 LiDAR sensor to acquire 3D point clouds of the human subject. This device provides high-precision depth information while inherently preserving user privacy by not capturing appearance textures. Moreover, it operates robustly under varying lighting conditions. In our setup, the LiDAR is positioned in front of the user, who can move within a range of 5–7 meters from the sensor.
Then, we employ the cutting-edge LiDAR-based framework LiveHPS++~\cite{ren2025livehps++}, which estimates pose parameters of human motions from scanned point clouds. More details are in supplementary material. 



\noindent\textbf{Robotic Interaction Model.} To enable robots to interact with humans naturally and smoothly, we have designed an efficient robot interaction model. This model uses motion-captured human movements as observational data to generate real-time motion responses, driving virtual robots to interact with real humans in 3D space. The technical approach is introduced in Section~\ref{sec:interactivemodel}.

\noindent\textbf{First-Person Perspective on AR Glasses.} We employ an AR device, i.e. PICO4 Ultra~\cite{PicoInteractive2022}, to visualize virtual robots from the first-person perspective. It communicates with the interactive model through TCP protocol to update the robot positions and poses. The virtual robot is seamlessly integrated into the real environment 
with photorealistic rendering.

\noindent\textbf{Third-Person Perspective on Screen.} To show the panoramic view of the interaction between the real human and the virtual robot, we set up an RGBD camera and calibrated its intrinsic and extrinsic parameters. After that, we render the virtual robot from the calibrated viewpoint and fuse it with the image captured by the camera based on their corresponding depth maps.


\subsection{Modules for Model Adaptation}

Model adaptation involves the collection of human-robot interaction records and human feedback, and utilizing them to effectively fine-tune the interactive model for robot skill enhancement.

\noindent\textbf{HRI Data and Feedback Collection.} Throughout the interactions process, we record the interaction data in different modalities, including the scanned point cloud and captured pose parameters of the human user, as well as the robot's pose parameters and skeleton keypoint positions. Meanwhile, users can provide feedback in various forms, such as direct ratings or detailed language descriptions of their interaction experience. 


\noindent\textbf{Model Fine-tuning.}
We classify the interaction data into positive and negative samples~\cite{zhu2025evolvinggrasp} based on the user ratings. Positive samples represent the cases that are well-received by users, while negative samples are those that users find unsatisfactory or unnatural. Our system supports a supervised learning paradigm to refine and adjust the interaction model by effective fine-tuning strategies. This allows the robot to better align with user needs and preferences. 

\section{Robotic Interaction Model}
\label{sec:interactivemodel}

Real-time, context-aware motion generation remains a persistent challenge for human-robot interaction, especially in dynamic and contact-rich scenarios.
Existing approaches~\cite{wang2021multi,xu2024regennet,mueller2024massively} for interactive robot motion generation often fail to adequately address real-time responsiveness and realistic interaction dynamics. To address these challenges, we propose a novel Robotic Interaction Model capable of generating coherent and timely interactive motion sequences for humanoid robot with three main technical advances:

\noindent\textbf{1) Affordance Predictor} precisely captures fine-grained spatial and geometric relationships between robot end-effectors and the human body, which is crucial for guiding the generation of precise and socially-appropriate interactive actions.
\noindent \textbf{2) Multi-Resolution Human Feature (MRHF) Learner} integrates human behavioral features at multiple temporal scales, improving the understanding of human intentions.
\noindent \textbf{3) Robot Rollout Loop} enables real-time, coherent, and context-adaptive robot motion, ensuring both responsiveness and long-term stability. It provides a solid guarantee for smooth and realistic interactive experience of the system.

\subsection{Model Input and Output}
Our model generates humanoid robot motions in an online fashion by conditioning on both human behavior and robot history. As shown in Fig.~\ref{fig:interactiveModel}, the model takes three inputs at each timestep.
\noindent\textbf{Textual Command ($T^C$)} is a user-specified action prompt such as \textit{``High-five''} or \textit{``Handshake''}, representing the intended interaction type. Our Action Embedding module uses a pre-trained CLIP~\cite{radford2021learning} text encoder to map it to a visual-aware feature $F^C$, which is then fused with motion features in downstream modules to guide context-aware motion generation.
\noindent\textbf{Historical Robot Motion ($J^{R}_{t-n\sim t}$)} is a sequence of past robot joint skeletons, stored in a rollout queue from timestep $t - n$ to $t$. \noindent\textbf{Human Motion Observation} includes human joint skeletons ($J^{H}_{t-n\sim t}$) and SMPL mesh vertices ($V^{H}_{t-n\sim t}$). Both are obtained by applying the SMPL model to the pose parameters \( P^{H}_{t-n\sim t} \) and translation parameters \( T^{H}_{t-n\sim t} \) from the motion capture mentioned in Section~\ref{sec:mocap}.

The model outputs a future sequence of predicted robot skeletons $J^R_{t+1 \sim t+k}$. We then employ a \textbf{receding horizon strategy}: only the immediate next frame $J^R_{t+1}$ is executed, while the rest of the sequence is discarded. This approach is crucial for balancing long-term motion coherence with real-time responsiveness. Predicting a multi-frame sequence ensures that the generated motion is smooth and natural, guided by sequence-level supervision. Conversely, executing only the first frame enables the robot to immediately adapt to dynamic human behavior by re-planning at every timestep. This continuous cycle not only guarantees high responsiveness but also enhances robustness by mitigating the accumulation of prediction errors.

\begin{table*}[]
\centering

\resizebox{\linewidth}{!}{ 
\tiny\begin{tabular}{c|c|cccccccccc}
\hline
Robot Type                      & Methods                                                & PA-MPJPE↓ & MPJPE↓ & Traj↓ & Orie↓ & C\_prec↑ & C\_rec↑ & C_acc↑& C_F1↑ & FID↓ & R-score↑ \\ \hline
\multirow{5}{*}{Unitree H1} & JRT~\cite{xu2023joint}          & 3.35& 11.99& 0.222& 36.51& 0.839& 0.798& 0.800& 0.818& 0.333& 0.420\\ 
\cline{2-12}& MRT~\cite{wang2021multi}         & 3.16 & 11.31 & 0.220 & 35.63 & \textbf{0.852} & 0.803 & 0.810 & 0.827 & 0.320 & 0.424 \\ 
\cline{2-12}& ReGenNet~\cite{xu2024regennet}   & 5.06 & 18.20 & 0.383 & 56.05 & 0.807 & 0.539 & 0.667 & 0.646 & 0.763 &0.363 \\ 
\cline{2-12}& SAST~\cite{mueller2024massively} & 5.31 & 19.77 & 0.322 & 46.69 & 0.805 & 0.742 & 0.753 & 0.772 & 0.527 & 0.409 \\ 
\cline{2-12}& Ours                             & \textbf{2.70} & \textbf{9.52} & \textbf{0.125} & \textbf{25.95} & 0.842 & \textbf{0.869} & \textbf{0.834} & \textbf{0.855} & \textbf{0.178} & \textbf{0.478} \\ \hline
\multirow{5}{*}{LEJU Kuavo} & JRT~\cite{xu2023joint}           & 3.11& 10.82& 0.199& 32.52& \textbf{0.849}& 0.788& 0.808& 0.817& 0.320& 0.445\\ 
\cline{2-12}& MRT~\cite{wang2021multi}         & 2.96 & 10.54 & 0.206 & 35.18 & 0.836 & 0.797 & 0.804 & 0.816 & 0.294 & 0.447 \\ 
\cline{2-12}& ReGenNet~\cite{xu2024regennet}   & 4.05 & 14.72 & 0.320 & 51.60 & 0.783 & 0.647 & 0.710 & 0.708 & 0.562 & 0.397 \\ 
\cline{2-12}& SAST~\cite{mueller2024massively} & 5.15 &  19.52 & 0.352 & 53.12  & 0.792 & 0.704  & 0.738 & 0.746 & 0.553 & 0.417 \\ 
\cline{2-12}& Ours                             & \textbf{2.46} & \textbf{8.74} & \textbf{0.122} & \textbf{24.77} & 0.844& \textbf{0.862} & \textbf{0.838} & \textbf{0.853} & \textbf{0.178} & \textbf{0.498} \\ \hline
\end{tabular}
}
\caption{{Quantitative comparison of human-robot interaction baselines on the Inter-HRI benchmark.} 
}
\vspace{-4ex}
\label{tab:hhi}
\end{table*}
\subsection{Affordance Predictor Module}

Effective 
interaction requires robots to understand where and how contact with a human should occur. Our Affordance Predictor learns a dense spatial field of fine-grained affordances—the Euclidean distance from every human body vertex to robot end-effectors (robot hands).

First, \textbf{Human Vertice Encoder} takes the human SMPL mesh sequence $V^H_{t-n\sim t}$ as input and extracts vertex-wise features $F^{V^H}_{t-n \sim t}$ using PointNet~\cite{qi2017pointnet}.
Second, \textbf{HRI Feature Fusion} uses an encoder-decoder Transformer to integrate the human vertex features $F^{V^H}_{t-n \sim t}$, the robot's historical joint sequence $J^R_{t-n\sim t}$, and the action embedding $F^C$ together for obtaining $F'^{V^H}_{t+1\sim t+k}$. Two cross-attention branches separately process motion and semantic cues. Finally, \textbf{Affordance Decoder} is followed to to acquire a dense affordance field $A_{t+1\sim t+k}$ via a lightweight MLP. $A_{t+1\sim t+k}$ 
denotes the spatial distance between each robot hand and every human body vertex over future timesteps, enabling the robot to reason about socially acceptable and physically feasible contact points, such as targeting the palm in a handshake.



\subsection{MRHF Learner Module}
While the Affordance Predictor captures spatial alignment at the interaction moment, fluid collaboration also demands temporal reasoning across different behavioral time scales - from slow postural changes to quick hand movements. Our MRHF Learner uses a hierarchical encoder to extract multi-scale features from human skeleton sequences $J_{t-n\sim t}^{H}$.
Specifically, low-, middle-, and high-level features are sequentially extracted through stacked encoders, containing convolutional and transformer layers, to capture both local motion nuances and long-term behavioral intent. These multi-scale representations are concatenated to form the Multi-Resolution Human Feature (MRHF) $F^{H_{MR}}_{t-n\sim t}$.
The MRHF combines with the robot's motion data in the subsequent Robot Rollout Loop, helping the robot respond naturally to human actions — quickly adapting to sudden movements while maintaining long-term interaction goals. 

\subsection{Robot Rollout Loop Module}
Unlike previous offline motion generation methods, which compute entire motion sequences in advance, our model adopts a frame-by-frame rollout strategy to generate online responsive motions.

\noindent\textbf{MRHF-enhanced Encoder} 
This encoder encodes historical robot joint sequences \(J^R_{t-n \sim t}\) and multi-resolution human features \(F^{H_{MR}}_{t-n \sim t}\) to extract context-aware robot motion features $F^{R}_{t-n \sim t} $.


\noindent\textbf{Affordance-guided Predictor} 
This predictor guides the motion generation process to spatially align the robot's motions with human interactive intent. Formally, it fuses the encoded robot features \(F^{R}_{t-n \sim t}\) with the affordance predictions and action embedding:
\[
F^{R'}_{t+1 \sim t+k} = Predictor_{Affordance}(F^{R}_{t-n \sim t}, A_{t+1 \sim t+k},F^{C}).
\]
By this, the model gains enhanced spatial awareness and predictive capability, 
 markedly elevating the precision of the interaction.

\noindent\textbf{MRHF-enhanced Decoder}
Finally, the fused features \(F^{R'}_{t+1 \sim t+k}\) are input into the MRHF-enhanced Decoder 
\[
J^{R}_{t+1 \sim t+k} = Decoder_{MRHF}(F^{R'}_{t+1 \sim t+k}, F^{H_{MR}}_{t-n \sim t})
\]
to generate the future robot joint sequences. 
By applying cross-attention to both robot history and multi-resolution human cues, the generated robot motion is not only temporally coherent but also spatially aligned with evolving human actions.
Only the immediate next frame \(J^{R}_{t+1}\) is forwarded 
for execution and rollout queue updates, enabling real-time interactive feedback and seamless motion rollout.

\subsection{Robot Motion Retargeting Module}


To transform the predicted robot joint skeletons into robot-specific motion parameters, we employ a sophisticated Robot Motion Retargeting Module, which is fundamentally built upon the SMPL Solver from the state-of-the-art LiveHPS++ motion capture system [Ren et al. 2025]. The module takes the continuous robot skeleton data as input and employs an attention-based neural network to estimate the robot's joint angles. Its architecture is designed to preserve the naturalness and fluidity of the motion while respecting the robot's specific mechanical constraints. Specifically, a transformer-based spatial encoder first models inter-joint relationships, and a subsequent bidirectional GRU captures temporal dynamics. This integrated approach ensures the generated robot poses \(P^{R}_{t+1}\) are not only accurate but also smooth and physically plausible, which is crucial for responsive and realistic human-robot interactions.

\section{System Evaluation}
\begin{table*}[]
\centering

\resizebox{\linewidth}{!}{ 
\tiny\begin{tabular}{c|c|cccccccccc}
\hline
Setting                     & Finetune Scale                                               & PA-MPJPE↓ & MPJPE↓ & Traj↓ & Orie↓ & C\_prec↑ & C\_rec↑ & C_acc↑& C_F1↑ & FID↓ & R-score↑ \\ \hline
\multirow{4}{*}{Cross Domains} & Base        &   4.57        &       17.12    &    0.286     &   28.14  &  0.891   &     0.773      &    0.807     &     0.828    &     0.289     &     /      \\ \cline{2-12} & $\sim$100         &    \textcolor[rgb]{0.07,0.55,0.15}{4.44}       &    \textcolor[rgb]{0.07,0.55,0.15}{16.32}       &    \textcolor[rgb]{0.07,0.55,0.15}{0.280}      &   \textcolor[rgb]{0.07,0.55,0.15}{26.86}  &  \textcolor[rgb]{0.07,0.55,0.15}{ 0.908}  &     \textcolor[rgb]{0.07,0.55,0.15}{0.774}      &         \textcolor[rgb]{0.07,0.55,0.15}{0.817}   &    \textcolor[rgb]{0.07,0.55,0.15}{ 0.836 }     & \textcolor[rgb]{0.07,0.55,0.15}{0.241}   &   /   \\ \cline{2-12}& $\sim$1000         &   \textcolor[rgb]{0.07,0.55,0.15}{4.30}     &     \textcolor[rgb]{0.07,0.55,0.15}{15.67}      &     \textcolor[rgb]{0.07,0.55,0.15}{0.275}      &      \textcolor[rgb]{0.07,0.55,0.15}{25.38}     &    \textcolor[rgb]{0.07,0.55,0.15}{0.918}    &    \textcolor[rgb]{0.77,0.15,0.32}{0.754}     &      \textcolor[rgb]{0.07,0.55,0.15}{0.812}     &      \textcolor[rgb]{0.07,0.55,0.15}{0.828}     &    \textcolor[rgb]{0.07,0.55,0.15}{0.202}      &      /    \\ \cline{2-12}& $\sim$10000   &      \textcolor[rgb]{0.07,0.55,0.15}{\textbf{3.85}}    &      \textcolor[rgb]{0.07,0.55,0.15}{\textbf{14.04}}     &    \textcolor[rgb]{0.07,0.55,0.15}{\textbf{0.232}}    &    \textcolor[rgb]{0.07,0.55,0.15}{\textbf{21.78}}     &    \textcolor[rgb]{0.07,0.55,0.15}{\textbf{0.923}}   &    \textcolor[rgb]{0.07,0.55,0.15}{\textbf{0.805}}  &     \textcolor[rgb]{0.07,0.55,0.15}{\textbf{0.842}}     &   \textcolor[rgb]{0.07,0.55,0.15}{\textbf{0.860}}   &    \textcolor[rgb]{0.07,0.55,0.15}{\textbf{0.094}}    &     /     \\ \hline
\multirow{4}{*}{Cross Categories} & Base &   3.06        &       10.72    &     0.151     &   34.49  &  0.822   &     0.845      &    0.807     &     0.833    &     0.274      &     0.448  \\ \cline{2-12}& $\sim$100         &     \textcolor[rgb]{0.07,0.55,0.15}{3.00}      &      \textcolor[rgb]{0.07,0.55,0.15}{10.52}     &     \textcolor[rgb]{0.07,0.55,0.15}{0.150}     &   \textcolor[rgb]{0.07,0.55,0.15}{34.20}  &  \textcolor[rgb]{0.77,0.15,0.32}{0.817}   &      \textcolor[rgb]{0.07,0.55,0.15}{0.849}     &  \textcolor[rgb]{0.77,0.15,0.32}{ 0.806 }     &     \textcolor[rgb]{0.77,0.15,0.32}{0.833}    &    \textcolor[rgb]{0.07,0.55,0.15}{ 0.262}      &    \textcolor[rgb]{0.07,0.55,0.15}{0.451 }      \\ \cline{2-12}& $\sim$1000         &    \textcolor[rgb]{0.07,0.55,0.15}{2.88}    &       \textcolor[rgb]{0.07,0.55,0.15}{10.06}    &     \textcolor[rgb]{0.07,0.55,0.15}{ 0.143}     &    \textcolor[rgb]{0.07,0.55,0.15}{32.08}       &   \textcolor[rgb]{0.07,0.55,0.15}{ 0.826}    &    \textcolor[rgb]{0.07,0.55,0.15}{\textbf{0.850}}   &       \textcolor[rgb]{0.07,0.55,0.15}{0.813 }   &     \textcolor[rgb]{0.07,0.55,0.15}{0.838}      &    \textcolor[rgb]{0.07,0.55,0.15}{0.241 }      &    \textcolor[rgb]{0.07,0.55,0.15}{0.455}      \\ \cline{2-12}& $\sim$10000   &  \textcolor[rgb]{0.07,0.55,0.15}{\textbf{2.67}}     &    \textcolor[rgb]{0.07,0.55,0.15}{\textbf{9.34}}       &    \textcolor[rgb]{0.07,0.55,0.15}{\textbf{13.05}}     &    \textcolor[rgb]{0.07,0.55,0.15}{ \textbf{30.02}}      &     \textcolor[rgb]{0.07,0.55,0.15}{\textbf{0.841}}   &     \textcolor[rgb]{0.07,0.55,0.15}{0.847}   &      \textcolor[rgb]{0.07,0.55,0.15}{\textbf{0.822}}    &    \textcolor[rgb]{0.07,0.55,0.15}{\textbf{0.844}}   &    \textcolor[rgb]{0.07,0.55,0.15}{\textbf{0.208}}    &      \textcolor[rgb]{0.07,0.55,0.15}{\textbf{0.465}}    \\ \hline
\end{tabular}
}
\caption{Results of interaction skill enhancement across different settings with varying finetune scales. \textcolor[rgb]{0.77,0.15,0.32}{Red} indicates a decrease while \textcolor[rgb]{0.07,0.55,0.15}{Green} indicates an increase relative to Base. Due to extreme category imbalance in Inter-Human, R-score is not applicable in \textit{Cross-Domain} setting.}
\vspace{-4ex}
\label{tab:finetune}
\end{table*}
We conduct system evaluation by assessing whether it meets the design goals. Accordingly, our experiments focus on answering three questions, presented in Sections~\ref{eval1}, \ref{eval2}, \ref{eval3}, respectively.

A. Does the interactive model produce reasonable and plausible robot reactions that ensure natural and smooth interactions?

B. Does the system support real-time and realistic interaction experience that helps humans adapt to human-robot interactions?

C. Does the system facilitate data collection and efficient fine-tuning that continuously enhance robot interaction skills?

\subsection{Evaluation of Interactive Model}
\label{eval1}
\noindent\textbf{Datasets.}
The Ground Truth (GT) data for training and evaluating our model was generated from the large-scale human-human interaction dataset, Inter-X \cite{xu2024inter}, addressing the lack of a comparable human-robot dataset. The GT generation process involved treating one human's motion in an interaction pair as the input, while the second human's response was converted into the target robot motion using an \textbf{offline motion retargeting} pipeline. This pipeline employs the kinematics-based methodology from Human2Humanoid \cite{he2024learning} to map human movements to the specific constraints of our robot models (LEJU Kuavo \cite{lejukuavorobot} and Unitree H1 \cite{unitreeh1robot}) ensuring the resulting motion is physically plausible and serves as a high-quality supervision signal. This offline generation of GT is distinct from the online, learning-based \textbf{Robot Motion Retargeting Module (Sec. 5.5)}, which is trained using this data to perform real-time predictions. Through this procedure, we created the \textit{Inter-HRI} benchmark for our experiments, which retains the original train-test split from Inter-X and is downsampled to 10FPS to suit our motion capture setup. For a comprehensive walkthrough of the GT dataset generation, please refer to the \textbf{Supplementary Material, Section 3.1}.

\noindent\textbf{Evaluation Metrics.} 
For measuring action precision, MPJPE is the mean per joint position error and PA-MPJPE aligns predictions via Procrustes analysis (rotation, translation, scale) before computing error~\cite{deitke2020robothor}. Both are reported in centimeters. Traj measures root joint trajectory error (\textit{cm}), while Orie measures orientation error(\textit{deg}).
For measuring contact quality, we use dimensionless metrics: precision ($C_{\text{prec}}$), recall ($C_{\text{rec}}$), accuracy ($C_{\text{acc}}$), and F1 score ($C_{\text{F1}}$).
For measuring generative fidelity, FID calculates Fréchet distance between inception network features. R-score evaluates text-motion alignment via Euclidean distances between embeddings and Top-3 retrieval accuracy. 

\begin{figure*}[h]
\centering
\includegraphics[width=1.0\linewidth]{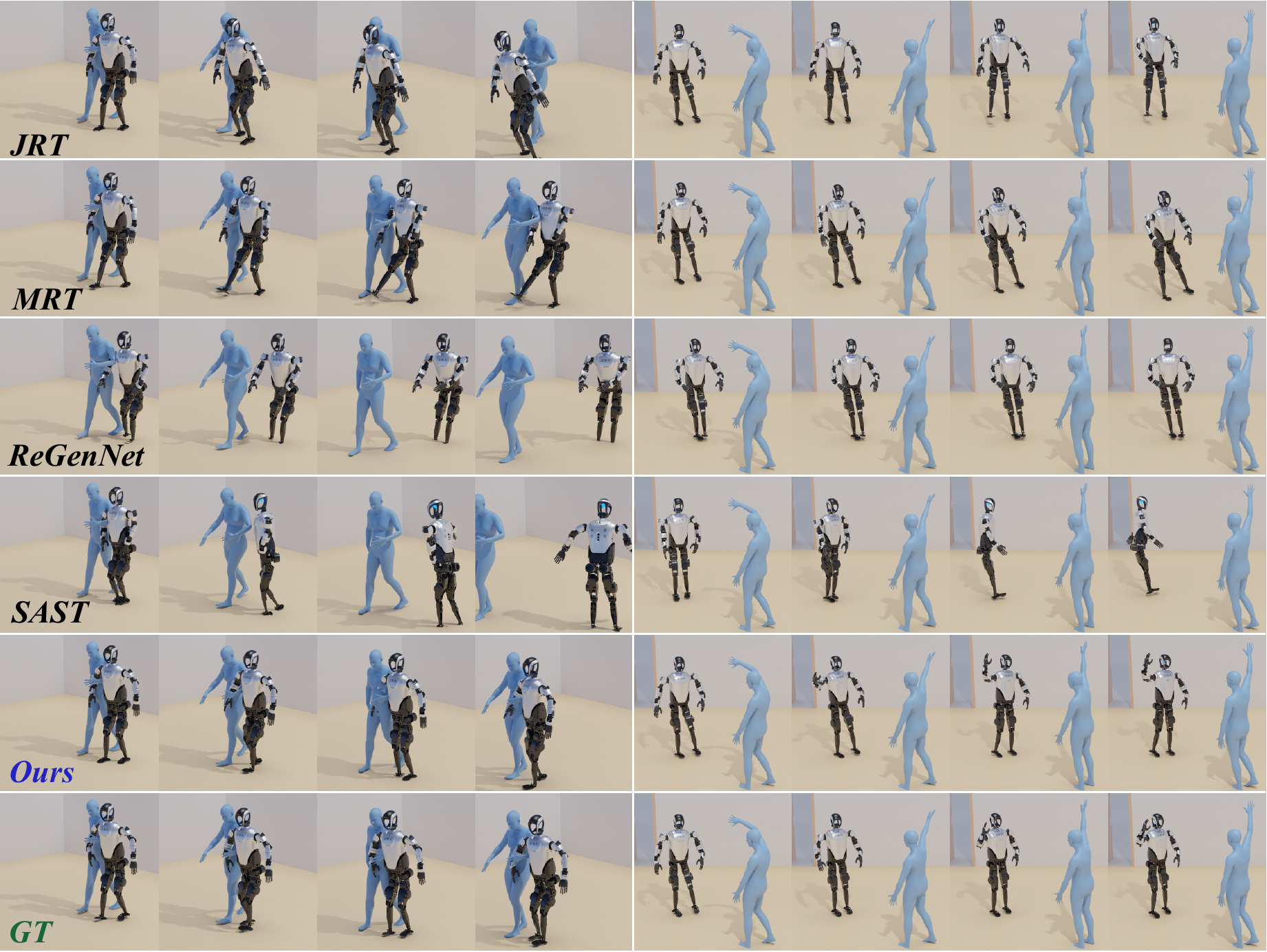}
\vspace{-20px}
\caption{Visualization comparison of robot motion generation for interactive sequences. We take two interaction cases as examples: "Link arms" (left) and "Wave" (right). Our method ("Ours") is shown alongside several baseline methods (JRT, MRT, ReGenNet, SAST) and the Ground Truth (GT). This visualization highlights that our model produces robot motions that are more natural, coherent, and closely aligned with the ground truth.}
\label{fig:visintermodel-comp}
\vspace{-5px}
\end{figure*}
\begin{figure*}[h]
\centering
\includegraphics[width=1.0\linewidth]{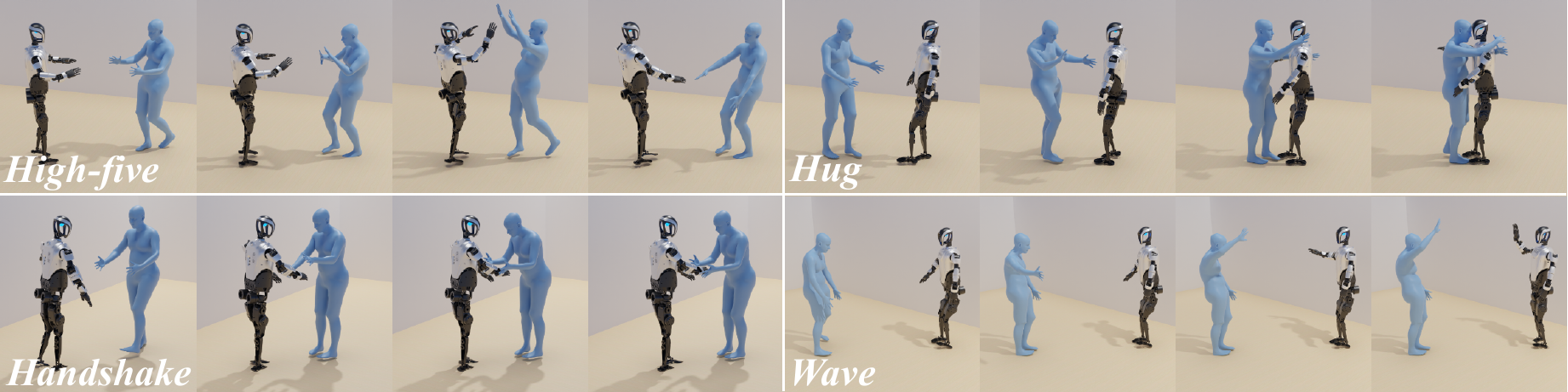}
\vspace{-20px}
\caption{Visualizations of diverse human-robot interactions generated by our proposed Robotic Interaction Model. These sequences illustrate the model's capability to produce contextually appropriate and varied robot responses to dynamic human actions.}
\label{fig:visintermodel-ours}
\vspace{-5px}
\end{figure*}

\noindent\textbf{Comparison.}
\label{sec:baseline}
We compare our Robotic Interaction Model against four recent state-of-the-art methods, all re-implemented or fine-tuned and evaluated fairly on the Inter-HRI benchmark.
As shown in Fig.~\ref{fig:visintermodel-comp} and Tab.~\ref{tab:hhi}, our model consistently outperforms these baselines with the lowest PA-MPJPE and MPJPE, and significant improvements in contact metrics ($C_{\text{prec}}$, $C_{\text{F1}}$), FID, and R-score, indicating superior spatial accuracy, physical realism, and semantic consistency.
These improvements result from our 
MRHF Learner, which captures hierarchical temporal behavior, and the Affordance Predictor, enabling precise, socially-aware contact reasoning (details in supplementary material). Together, they enable realistic, contextually adaptive, and real-time human-robot interactions. More our results of diverse interaction types are presented in Fig.~\ref{fig:visintermodel-ours}.

\noindent\textbf{Human Evaluation.}We also compared the quality of generated actions across ten representative action types in the Inter-X dataset using a 5-point Likert scale, where a higher score indicates a more natural and accurate result. 30 computer science students (15 male, 15 female) were asked to rate the outputs of five methods: JRT, MRT, ReGenNet, SAST, and our proposed approach. The results show that our method significantly outperforms the baselines, achieving an average score of 4.5, while the closest competitor, JRT, received an average of 2.46. MRT and ReGenNet followed with average scores of 2.1 and 1.86, respectively, and SAST ranked lowest with 1.64. These findings suggest that our approach generates actions that are perceived by humans as more realistic and semantically appropriate than those produced by existing methods.

\subsection{Evaluation of Real-Time and Realistic HRI}
\label{eval2}

\noindent\textbf{Time Performance.} Our system involves a 10FPS LiDAR device, a host server running the motion capture and interactive model, and display ends, either an AR glass or a notebook for different perspectives.
All modules run in parallel with cached queues between them.
On a host server with an i9-14900kf CPU and a Nvidia A6000 GPU, it achieves a speed of 40FPS with overall latency less than 0.2sec.

\noindent\textbf{Quality Performance.} 
In our system, the Ouster-1-128 LiDAR exhibits a scanning error of $\pm$1-3$\,cm$ in indoor environments. For motion capture, we employ a noise-resistant algorithm, which achieves a human keypoint estimation error of $6.2~cm$ and a joint angle error of $15.40~degree$ (evaluated on the FreeMotion 
dataset). These errors define the input data error for the interaction model. On the other hand, 
between the robot predicted by the interactive model and that presented to the user, i.e. input and output of the retargeting module, the joint angle error is $3.80~degree$ for LEJU Kuavo and $6.81~degree$ for Unitree H1 in the Inter-HRI dataset.
\color{black}

\noindent\textbf{User Empirical Test.} To evaluate our system, we recruited a total of 50 participants (33 male, 17 female), aged between 18 and 56 years and ranging in height from 162 cm to 191 cm, with diverse academic and professional backgrounds in robotics, computer graphics, and augmented reality. Each participant required to test 5 types of interaction 3 iterations per type. 
After the interactions, the user rates their interaction experience on a 5-Likert scale from different dimensions, where 1 indicates "poor" and 5 indicates "excellent".
Among the dimensions, 
\emph{interaction completion} reflects how well the user considers the interaction as successfully completed. \emph{Realism} is whether the user perceives a realistic visualization and a vivid interaction with the virtual robot. \emph{Rationality} and 
\emph{fidelity} measure whether the robot's reaction can interact with the human user appropriately and whether it acts like a physical humanoid robot. \emph{Real-time performance} is whether the user considers the robot's reaction as smooth and timely. 
\emph{Usability} and \emph{willing-to-use} reflects whether the users view the system as easy-to-use and would like to continue performing the human-robot interaction via our system.

\begin{figure*}[b]
\centering
\includegraphics[width=1.0\linewidth]{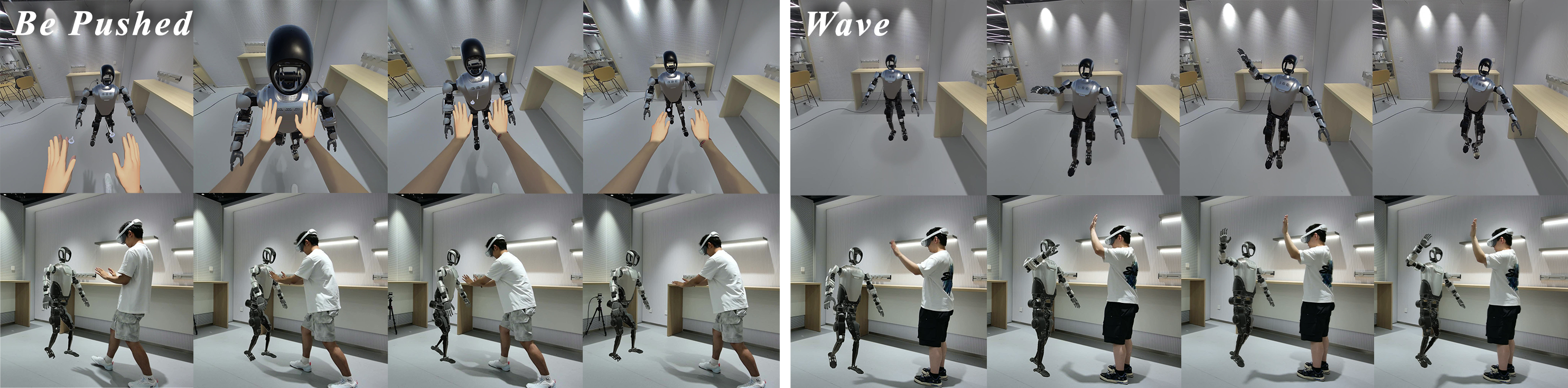}
\vspace{-20px}
\caption{The human-robot interactions presented in our system. Top: The first-person perspective visualized via the AR glasses. Bottom: The third-person perspective produced by fusing the captured human user in the physical environment and the rendered virtual robot.}
\label{fig:systeminter}
\end{figure*}

\begin{figure*}[h]
\centering
\includegraphics[width=1.0\linewidth]{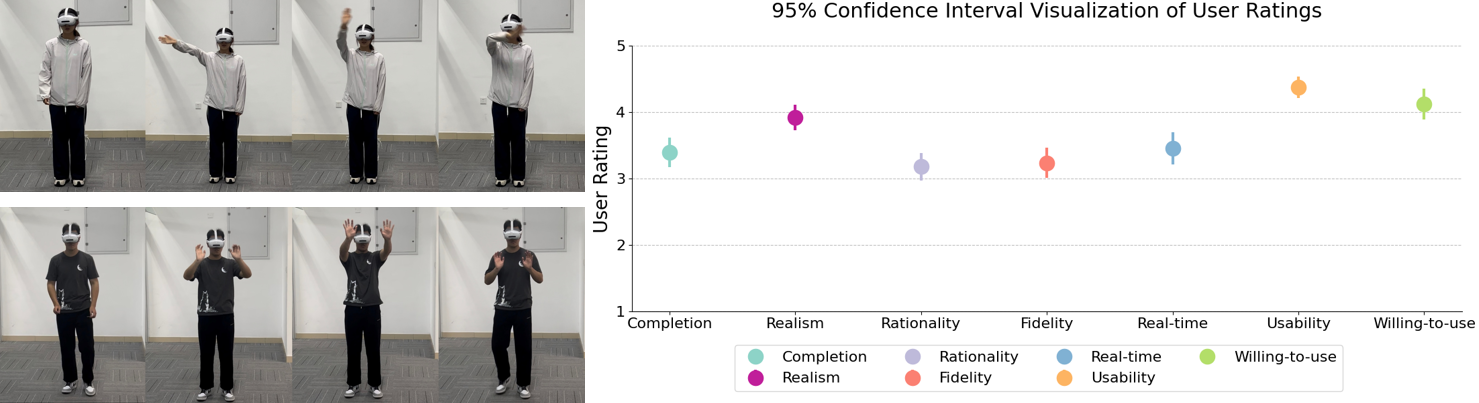}
\vspace{-20px}
\caption{Process and results of the user empirical test. Left: Participants interacting with virtual robots via our system. Right: The 95$\%$ confidence interval visualization of user ratings showing the mean and boundary of the confidence intervals. Here, 1 indicates "poor" and 5 indicates "excellent".}
\label{fig:userstudy}
\end{figure*}

The process and statistics are presented in Fig.~\ref{fig:systeminter} and~\ref{fig:userstudy}. All dimensions receive average ratings above the moderate level. Our system exhibits high realism and moderate real-time characteristics, as well as positive rationality and fidelity, yet distinguishable from real human-human interactions. More importantly, most users consider our system as quite easy-to-use and would like to continue using it. 

\begin{figure*}[b]
\centering
\includegraphics[width=1.0\linewidth]{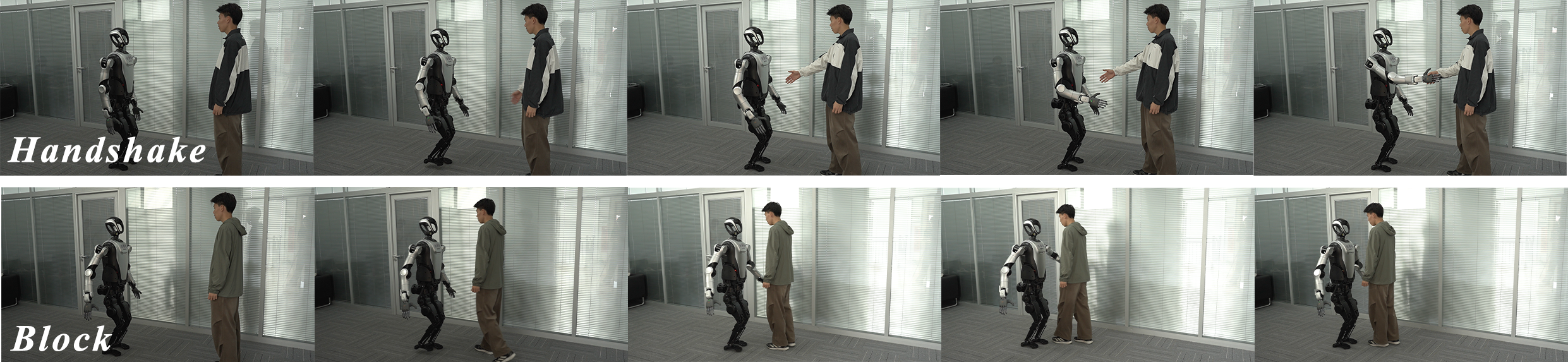}
\vspace{-20px}
\caption{A real LEJU Kuavo robot performing the robot movements generated by the robotic interaction model. Since the real robot exhibits a high consistency with the appearance of the virtual robot shown in our system, a human user can easily and smoothly interact with the real robot.}
\vspace{-20px}
\label{fig:realrobot}
\end{figure*}

\noindent\textbf{Real Robot Test.} 
We deploy the virtual robot reactions to a real LEJU Kuavo robot. That is, we preserve the exact upper body movements and body center trajectory, and slightly tune the lower body movements with its default offline simulator software to maintain the robot's balance. As in Fig. ~\ref{fig:realrobot}, the appearance and movement of the real robot exhibits a high consistency with that presented by our system. This test indicates that the human-robot interaction experience via our system is reliable enough to assist human users in getting used to interacting with robots in the real world.

\noindent\textbf{Discussion.} Our system currently supports a single active user, although the motion capture module can track multiple individuals. For reliable operation, users must remain within the LiDAR’s line of sight and unobstructed by obstacles. If no human is detected, the prediction pipeline pauses and the robot remains visible but stationary in the AR interface, thereby preventing unintended behaviors while maintaining scene continuity.

\subsection{Evaluation of Robot Interaction Skill Enhancement}
\label{eval3}

\noindent\textbf{Setup.} We conduct model adaptation with a base model trained on a subset of Inter-HRI with 20 basic actions. The fine-tuning uses the loss function leveraging both positive and negative samples 
\[
L = L_{\text{pos}} - \alpha L_{\text{neg}},
\]
where $L_{\text{pos}}$  and $L_{\text{neg}}$ are reconstruction loss terms to encourage learning from desirable interactions while penalizing undesired behaviors, and $\alpha$ controls the negative loss influence.

We evaluate two settings by processing existing datasets to mimic the interactions collected with our system. \textit{Cross Categories} focuses on interactions outside the 20 base types. \textit{Cross Domain} simulate real-world user movements by filtering samples of these types from another interaction dataset InterHuman~\cite{liang2024intergen}. 
In both settings, we use 1:1 positive and negative samples. Negative samples are produced by (1) injecting random noise to robot motions, (2) making static sequences like robot not responding to humans, and (3) re-pairing human-robot movements like mismatched interactions.

\noindent\textbf{Results.}
Tab.~\ref{tab:finetune} presents the quantitative results for \emph{Cross Categories} and \emph{Cross Domain}.
Increasing data scale (to 100, 1,000, 10,000 samples) consistently improves performance on both settings. These improvements confirm that our model adaptation mechanism can effectively utilize
various human-robot interaction data collected with our system to enhance robot interaction skills.

\section{Conclusion and Future work}
\label{sec:conclusion}

In this paper, we propose a novel human-in-the-loop cyber-physical robotic interaction system, SymBridge, by utilizing AR technology to deliver a highly realistic human-robot interaction experience. 
This system allows for diverse testing scenarios and human feedback collection, enabling model adjustments and robot evolution. This facilitates mutual adaptation between humans and robots, driving human-robot symbiosis in a safe, efficient, and cost-effective way.

In the future, we will keep enhancing our interactive system towards human-robot symbiosis. First, multi-modal sensors such as portable haptic sensors can be integrated to improve realism of the interaction process. Second, we intend to develop reinforcement learning-based online fine-tuning methods to rapidly adjust the robot behavior based on real-time human feedback. Moreover, we will create more robust and generalized human-robot interaction and collaboration base models, further extending the application scenarios and accelerating research progress in related fields.

\section{Acknowledgement}
This work is supported by the National Natural Science Foundation of China (U23A20312, No.62302269, No.62206173), Shandong Provincial Natural Science Foundation (No.ZR2023QF077), Shandong Province Excellent Young Scientists Fund Program (Overseas) (No.2023HWYQ-034), Shanghai Frontiers Science Center of Human-centered Artificial Intelligence (ShangHAI), MoE Key Laboratory of Intelligent Perception and Human-Machine Collaboration (KLIP-HuMaCo).

\bibliographystyle{formats/ACM-Reference-Format}
\bibliography{Reference} 

\newpage

\begin{figure*}[b]
\centering
\includegraphics[width=1.0\linewidth]{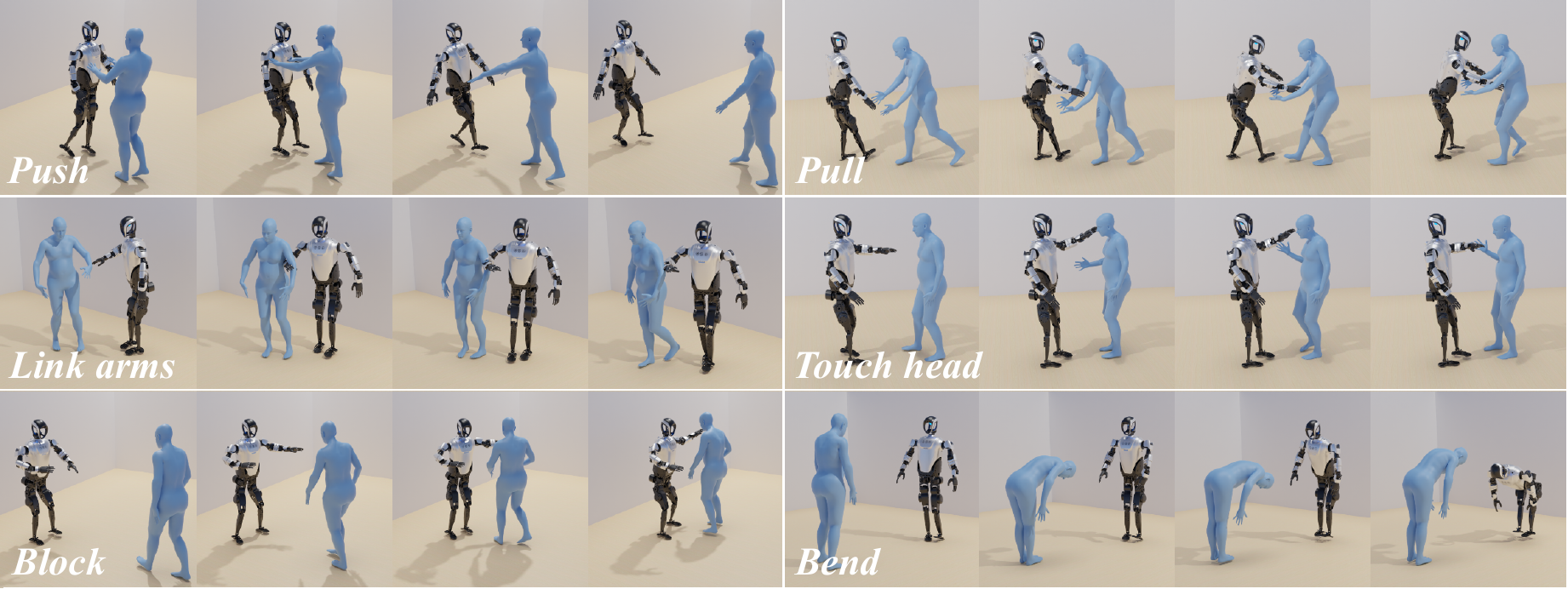}
\end{figure*}

\vspace{-20px}

\appendix

\end{document}